\documentclass[12pt,a4]{article}
\usepackage[dvips]{color,graphicx}
\usepackage{bm,psfrag,afterpage,color,amsmath,amssymb,latexsym,amsthm,url,lscape}

\renewcommand{\arraystretch}{0.8}

\renewcommand{\arraystretch}{1.0}

\newcommand{\argmin}{\mathop{\rm arg~min}\limits}

\usepackage[subrefformat=parens]{subcaption}
\usepackage[subrefformat=parens]{caption}

\makeatletter
\def\eqnarray{\stepcounter {equation}\let \@currentlabel =\theequation
\global \@eqnswtrue
\global \@eqcnt \z@ \tabskip \@centering \let \\=\@eqncr
$$\halign to \displaywidth \bgroup \@eqnsel \hskip \@centering
$\displaystyle \tabskip \z@ {##}$&\global \@eqcnt \@ne \hfil
${\mbox{}##\mbox{}}$\hfil &\global \@eqcnt \tw@
$\displaystyle \tabskip \z@ {##}$\hfil \tabskip \@centering
&\llap {##}\tabskip \z@ \cr}
\makeatother
\renewcommand{\arraystretch}{0.9}

\begin{document}

{\baselineskip = 8mm 

\begin{center}
\textbf{\LARGE Sparse principal component regression via singular value decomposition approach} %\\[1mm]
\end{center}
%Predictive model selection criteria for Bayesian lasso
%Model selection criteria for Bayesian lasso from predictive viewpoints

\begin{center}
{\large Shuichi Kawano$^{1}$}
\end{center}

\begin{center}
\begin{minipage}{14cm}
{
\begin{center}
{\it {\footnotesize 

\vspace{1.2mm}

%Department of Social Intelligence and Informatics, \\

$^1$Graduate School of Informatics and Engineering, The University of Electro-Communications, 
1-5-1 Chofugaoka, Chofu-shi, Tokyo 182-8585, Japan. \\

\vspace{1.2mm}

%$^2$ The Institute of Statistical Mathematics, \\
%10-3 Midori-cho, Tachikawa, Tokyo 190-8562, Japan. \\
%
%\vspace{1.2mm}
%
%$^3$ Department of Statistical Science, The Graduate University for Advanced Studies, \\
%10-3 Midori-cho, Tachikawa, Tokyo 190-8562, Japan. \\
%
%\vspace{1.2mm}
%
%$^4$ Department of Mathematical Statistics, Graduate School of Medicine, Nagoya University, 
%65 Tsurumai-cho, Showa-ku, Nagoya 466-8550, Japan. \\
%
%\vspace{1.2mm}
%
%$^5$ Mammalian Genetics Laboratory, 
%National Institute of Genetics, \\ Mishima, Shizuoka 411-8540, Japan.\\
%
%\vspace{1.2mm}

%$^4$ Transdisciplinary Research Integration Center, \\
%Research Organization of Information and Systems, Minato-ku, Tokyo 105-0001, Japan. \\
%}}
}}

\vspace{2mm}

skawano@ai.lab.uec.ac.jp

\end{center}

%-----------------------------------------------
%\vspace{1mm} 

%{\small {\bf Abstract:} \
%Principal component regression (PCR) is a widely-used two-stage procedure: we first perform principal component analysis (PCA) and next consider a regression model in which selected principal components are regarded as new explanatory variables. 
%We should remark that PCA is based only on the explanatory variables, so the principal components are not selected using the information on the response variable. 
%In this paper, we propose a one-stage procedure for PCR in the framework of generalized linear models. 
%The basic loss function is based on a combination of the regression loss and PCA loss. 
%The estimate of the regression parameter is obtained as the minimizer of the basic loss function with sparse penalty. 
%The proposed method is called the sparse principal component regression for generalized linear models (SPCR-glm). 
%SPCR-glm enables us to obtain sparse principal component loadings that are related to a response variable, because the two loss functions are simultaneously taken into consideration. 
%A combination of loss functions may cause the identifiability problem on parameters, but it is overcome by virtue of sparse penalty. 
%The sparse penalty plays two roles in this method. 
%The parameter estimation procedure is proposed using various update algorithms with the coordinate descent algorithm. 
%}

}
\end{minipage}
\end{center}

\vspace{1mm} 

\begin{abstract}
\noindent 
Principal component regression (PCR) is a two-stage procedure: the first stage performs principal component analysis (PCA) and the second stage constructs a regression model whose explanatory variables are replaced by principal components obtained by the first stage. 
Since PCA is performed by using only explanatory variables, the principal components have no information about the response variable. 
To address the problem, we propose a one-stage procedure for PCR in terms of singular value decomposition approach. 
Our approach is based upon two loss functions, a regression loss and a PCA loss, with sparse regularization. 
The proposed method enables us to obtain principal component loadings that possess information about both explanatory variables and a response variable. 
An estimation algorithm is developed by using alternating direction method of multipliers. 
We conduct numerical studies to show the effectiveness of the proposed method.

%Principal component regression (PCR) is a widely used two-stage procedure: principal component analysis (PCA), followed by regression in which the selected principal components are regarded as new explanatory variables in the model. 
%Note that PCA is based only on the explanatory variables, so the principal components are not selected using the information on the response variable. 
%In this paper, we propose a one-stage procedure for PCR in the framework of generalized linear models. 
%The basic loss function is based on a combination of the regression loss and PCA loss. 
%An estimate of the regression parameter is obtained as the minimizer of the basic loss function with a sparse penalty. 
%We call the proposed method sparse principal component regression for generalized linear models (SPCR-glm). 
%Taking the two loss function into consideration simultaneously, SPCR-glm enables us to obtain sparse principal component loadings that are related to a response variable. 
%However, a combination of loss functions may cause a parameter identification problem, but this potential problem is avoided by virtue of the sparse penalty. 
%Thus, the sparse penalty plays two roles in this method. 
%The parameter estimation procedure is proposed using various update algorithms with the coordinate descent algorithm. 
%We apply SPCR-glm to two real datasets, doctor visits data and mouse consomic strain data. 
%SPCR-glm provides more easily interpretable principal component (PC) scores and clearer classification on PC plots than the usual PCA. 
\end{abstract}

\begin{center}
\begin{minipage}{14cm}
{
\vspace{3mm}


%\vspace{3mm}

%{\small \noindent {\bf Mathematics Subject Classication:} 62H30, 62G05, 68T10.}
}
\end{minipage}
\end{center}

%%%%%%%%%%%%%%%%%%%%%%%%%%%%%%%%%%%%%%%%%%%%%%%%%%%%%%%%%%%%%%%%%%%%%%%%%%%%%%%%%%%%%%%%%%%%%%%%%%%%%%%%%%%%%%%%%%%%%%%%%%%%%%%%%%%%%%%%%%%%%%%%%%%%%%%%%%%%%%%%
\baselineskip = 8mm

%\vspace{8mm}

\section{Introduction}

%\cite{bastien2005pls}
Principal component regression (PCR), invented by Massy (1965) and Jolliffe (1982), is widely used in various fields of research including chemometrics, bioinformatics, and psychology, and then has been extensively studied by a lot of researchers (Frank and Friedman, 1993; Hartnett \textit{et al.}, 1998; Rosital \textit{et al.}, 2001; Reiss and Ogden, 2007; Wang and Abbott, 2008; Chang and Yang, 2012; Febrero-Bande \textit{et al.}, 2017; Dicker \textit{et al.}, 2017).  
PCR is based on a two-stage procedure: one performs principal component analysis (PCA) (Pearson, 1901; Jolliffe, 2002), followed by regression in which explanatory variables are the selected principal components. 
However, owing to the two-stage procedure, the principal components do not have information on the response variable. 
This causes low prediction accuracy for PCR, if the response variable is related with the principal components having small eigenvalues.

To address the problem, a one-stage procedure for PCR has been proposed by Kawano \textit{et al.} (2015). 
Its one-stage procedure is developed by combining a regression squared loss function with a sparse PCA (SPCA) loss function by Zou \textit{et al.} (2006). 
The estimate of the regression parameter and loading matrix in PCA is obtained as the minimizer of the combination of two loss functions with sparse regularization. 
By virtue of sparse regularization, it enables us to obtain sparse estimates of the parameters. 
Kawano \textit{et al.} (2015) called the one-stage procedure sparse principal component regression (SPCR). 
Kawano \textit{et al.} (2018) have also extended SPCR in the framework of generalized linear models. 
It is, however, doubtful whether using the PCA loss function by Zou \textit{et al.} (2006) is the best choice for SPCR, because there exist various formulae for PCA.

This paper proposes a novel formulation for SPCR. 
As a PCA loss for SPCR, we adopt a loss function by the singular value decomposition approach (Shen and Huang, 2008). 
Using the basic loss function, a combination of the PCA loss and the regression squared loss, with sparse regularization, we derive an alternative formulation for SPCR. 
We call the proposed method sparse principal component regression based on singular value decomposition approach (SPCRsvd). 
An estimation algorithm of SPCRsvd is developed by using an alternating direction method of multipliers (Boyd \textit{et al}., 2010) and a linearized alternating direction method of multipliers (Wang and Yuan, 2012; Li \textit{et al}., 2014). 
%We show the effectiveness of SPCRsvd through numerical studies. 
%The performance of SPCRsvd is competitive or better than that of SPCR. 

The rest of this paper is organized as follows. 
In Section \ref{sec:preliminaries}, we review SPCA by Zou \textit{et al.} (2006) and Shen and Huang (2008), and SPCR by Kawano \textit{et al.} (2015). 
We present SPCRsvd in Section \ref{sec:SPCRsvd}. 
Section \ref{sec:implementation} derives two computational algorithms for SPCRsvd and discusses the selection of tuning parameters included in SPCRsvd. 
Monte Carlo simulations and real data analyses are presented in Section \ref{sec:NumericalStudy}. 
Conclusions are given in Section \ref{sec:Conclusions}. 
%Supplementary materials can be found at \url{https://github.com/ShuichiKawano/spcr-svd/blob/master/suppl_spcr-svd.pdf}. 
Supplementary materials can be found at https:\slash\slash{}github.com\slash{}ShuichiKawano\slash{}spcr-svd\slash{}blob\slash{}master\slash{}suppl\_spcr-svd.pdf. 

%The R language software package {\tt spcr}, which implements SPCR-glm, is available on the Comprehensive R Archive Network (R Core Team, 2016). 

%%%%%%%%%%%%%%%%%%%%%%%%%%%%%%%%%%%%%%%%%%%%%%%%%%%%%%%%%%%%%%%%%%%%%%%%%%%%%%%%%%%%%%%%%%%%%%%%%%%%%%%%%%%%%%%%%%%%%%%%%%%%%%%%%%%%%%%%%%%%%%%%%%%%%%%%%%%%%%%%
\section{Preliminaries}
\label{sec:preliminaries}

%%% Write SPCA by Zou et al. and by Shen and Huang.
%%% Write SPCR by Kawano et al. (2015)

%%%%%%%%%%%%%%%%%%%%%%%%%%%%%%%%%%%%%%%%%%%%%%%%%%%%%%%%%%%%%%%%%%%%%%%%%
\subsection{Sparse principal component analysis}
\label{sec:SPCA}
PCA finds a loading matrix that induces a low-dimensional structure in data.  %linear combination of the variables such that the variance of data projected on the linear combination is maximized. 
To interpret the principal component loading matrix easily, SPCA has been proposed. 
Many researchers have studied various formulae for SPCA until now (Zou \textit{et al.}, 2006; d'Aspremont \textit{et al.}, 2007; Shen and Huang, 2008; Witten \textit{et al.}, 2009; Vu \textit{et al.}, 2013; Bresler \textit{et al.}, 2018; Chen \textit{et al.}, 2019; Erichson \textit{et al.}, 2019).
For overview of SPCA, we refer the reader to Zou and Xue (2018) and references therein.
In this subsection, we review two formulae for SPCA by Zou \textit{et al.} (2006) and Shen and Huang (2008).

Let $X=({\bm x}_1, \ldots, {\bm x}_n)^T$ denote an $n \times p$ data matrix, where $n$ and $p$ are the number of observations and the number of variables, respectively. 
Without loss of generality, we assume that the columns of the matrix $X$ are centered. 
Zou \textit{et al.} (2006) proposed SPCA by
\begin{equation}
\min_{A, B}  \left\{ \sum_{i=1}^n \| {\bm x}_i - A B^T {\bm x}_i \|^2_2 + \lambda \sum_{j=1}^k \| {\bm \beta}_j \|^2_2 + \sum_{j=1}^k \lambda_{1,j} \| {\bm \beta}_j \|_1 \right\} \ \ {\rm subject \ to} \ \  A^T A = I_{k}, 
\label{eq:spca}
\end{equation}
where $A$ and $B=({\bm \beta}_1,\ldots,{\bm \beta}_k)$ are $p \times k$ principal component (PC) loading matrices, $k$ denotes the number of principal components, $I_k$ is the $k \times k$ identity matrix, $\lambda,\lambda_{1,1},\ldots,\lambda_{1,k}$ are regularization parameters with non-negative value, and $\| \cdot \|_q$ is the $L_q$ norm for an arbitrary finite vector. 
The SPCA formulation can be regarded as a least squares approach. 
The first term represents to perform PCA by least squares. 
The second and third terms represent sparse regularization similar with the elastic net penalty (Zou and Hastie, 2005). 
The terms enables us to set some estimates of $B$ to zero. 
If $\lambda=0$, the regularization terms reduce to the adaptive lasso penalty (Zou, 2006). %, and then if $\lambda_{1,1}=\cdots=\lambda_{1,k}$ in addition to $\lambda=0$, to the lasso penalty (Tibshirani, 1996). 

A simple calculation leads to 
\begin{equation}
\min_{A, B} \sum_{j=1}^k  \left\{ \| X {\bm \alpha}_j - X {\bm \beta}_j  \|^2_2 + \lambda \| {\bm \beta}_j \|^2_2 +  \lambda_{1,j} \| {\bm \beta}_j \|_1 \right\} \ \ {\rm subject \ to} \ \  A^T A = I_{k}. 
\label{eq:spca2}
\end{equation}
This minimization problem is easy to optimize the parameters $A$ and $B$. 
Given a fixed $A$, the SPCA problem \eqref{eq:spca2} turns out to be a simple elastic net problem. 
Therefore, the estimate of $B$ can be obtained by the least angle regression algorithm (Efron {\it et al.}, 2004) or the coordinate descent algorithm (Friedman \textit{et al.}, 2007; Wu and Lange, 2008). 
Given a fixed $B$, the estimate of $A$ is obtained by solving the reduced rank Procrustes rotation problem (Zou \textit{et al.}, 2006). 
By alternating the procedures, we obtain the final estimates $\hat{A}$ and $\hat{B}$ of $A$ and $B$, respectively. 
Note that only $\hat{B}$ is used as the principal component loading matrix in Zou \textit{et al.} (2006).

On the other hand, Shen and Huang (2008) proposed another formulation of SPCA, which can be regarded as a singular value decomposition (SVD) approach. 
Consider a low rank approximation of the data matrix $X$ by SVD in the form
\begin{equation}
UDV^T = \sum_{k=1}^r d_k {\bm u}_k {\bm v}_k^T,
\label{eq:SVD}
\end{equation}
where $U=({\bm u}_1,\ldots,{\bm u}_r)$ is an $n \times r$ matrix with $U^T U = I_r$, $V=({\bm v}_1,\ldots,{\bm v}_r)$ is an $r \times r$ orthogonal matrix, $D = {\rm diag} (d_1,\ldots,d_r)$, and $r < \min(n,p)$. 
The singular values are assumed to be ordered such that $d_r \geq \cdots \geq d_p \geq 0$. 
By the connection between PCA and SVD, Shen and Huang (2008) obtained the sparse PC loading by estimating $V$ with sparse regularization.

To achieve sparseness of $V$, Shen and Huang (2008) adopted the rank-one approximation procedure. 
First we obtain the first PC loading vector $\tilde{\bm v}_1$ by solving the minimization problem
\begin{equation}
\min_{\tilde{\bm u}_1, \tilde{\bm v}_1} \left\{ \| X - \tilde{\bm u}_1 \tilde{\bm v}_1^T \|_F^2 + \lambda P (\tilde{\bm v}_1) \right\} \ \ {\rm subject \ to} \ \ \| \tilde{\bm u}_1 \|_2=1.
\label{eq:spcaSVD_first}
\end{equation}
Here $\tilde{\bm u}_1, \tilde{\bm v}_1$ are defined as rescaled vectors such that $\tilde{\bm u}_1 \tilde{\bm v}_1^T= d_1 {\bm u}_1 {\bm v}_1^T$, $P(\cdot)$ is a penalty function that induces the sparsity of $\tilde{\bm v}_1$, and $\| \cdot \|_F$ is the Frobenius norm defined by $\| A \|_F = \sqrt{ {\rm tr}(A^T A) }$ for an arbitrary matrix $A$. 
As the penalty function, Shen and Huang (2008) used the lasso penalty (Tibshirani, 1996), the hard-thresholding penalty (Donoho and Johnstone, 1994), and the smoothly clipped absolute deviation (SCAD) penalty (Fan and Li, 2001). 
It is easy to solve the rank-one approximation problem \eqref{eq:spcaSVD_first}; see Algorithm 1 of Shen and Huang (2008). 
The remaining PC loading vectors are provided by performing the rank-one approximations of the corresponding residual matrices. 
For example, to derive the second PC loading vector $\tilde{\bm v}_2$, we solve the minimization problem
\begin{equation*}
\min_{\tilde{\bm u}_2, \tilde{\bm v}_2} \left\{ \| X^\dagger - \tilde{\bm u}_2 \tilde{\bm v}_2^T \|_F^2 + \lambda P (\tilde{\bm v}_2) \right\} \ \ \ {\rm subject \ to} \ \ \ \| \tilde{\bm u}_2 \|_2=1,
\end{equation*}
where $X^\dagger = X - \tilde{\bm u}_1 \tilde{\bm v}_1^T$. 
The regularization parameter $\lambda$ is selected by cross-validation.

%%%%%%%%%%%%%%%%%%%%%%%%%%%%%%%%%%%%%%%%%%%%%%%%%%%%%%%%%%%%%%%%%%%%%%%%%
\subsection{Sparse principal component regression}
\label{sec:SPCR}

For a one-dimensional continuous response variable $Y$ and a $p$-dimensional explanatory variable $\bm x$, we postulate to obtain a dataset $\{ (y_i, {\bm x}_i) ; i=1,\ldots,n \}$. 
We assume that the response variable is explained by variables composed by PCA of $X = ({\bm x}_1,\ldots, {\bm x}_n)^T$.
Ordinary PCR is a regression model with a few PC scores corresponding to large eigenvalues. 
Note that the PC scores are previously constructed by PCA. 
This two-stage procedure might then fail to predict the response if the response variable is related with PCs corresponding to small eigenvalues.

To attain the one-stage procedure for PCR, Kawano \textit{et al.} (2015) proposed SPCR that is formulated by the following minimization problem
\begin{align}
& \min_{A, B, \gamma_0, {\bm \gamma}} \Bigg\{ \sum_{i=1}^n \left(y_i - \gamma_0 - {\bm \gamma}^T B^T {\bm x}_i \right)^2 + w \sum_{i=1}^n \| {\bm x}_i - A B^T {\bm x}_i \|^2_2   \nonumber \\
&  \hspace{20mm} + \lambda_{\beta} \xi \sum_{j=1}^k \| {\bm \beta}_j \|^2_2 + \lambda_{\beta} (1-\xi) \sum_{j=1}^k  \| {\bm \beta}_{j} \|_1 + \lambda_{\gamma} \| {\bm \gamma} \|_1\Bigg\}  \label{eq:spcr} \\
& {\rm subject \ to} \ \ \ A^T A = I_{k}, \nonumber
\end{align}
 where $\gamma_0$ is an intercept, ${\bm \gamma} = (\gamma_1,\ldots,\gamma_k)^T$ is coefficients for regression, $ \lambda_{\beta}$ and $\lambda_{\gamma}$ are regularization parameters with non-negative values, $w$ is a tuning parameter with non-negative value, and $\xi$ is a tuning parameter in $[0,1]$. 
The first term in Formula (\ref{eq:spcr}) is the least squared loss function including the PCs $B^T {\bm x}$ as explanatory variables, while the second term is the PCA loss function used in SPCA by Zou \textit{et al.} (2006). 
Sparse regularization in SPCR has two roles: sparseness and identifiability of parameters. 
For the identifiability by sparse regularization, we refer to Jennrich (2006), Choi \textit{et al.} (2011), and Kawano \textit{et al.} (2015). 
Kawano \textit{et al.} (2018) also extended SPCR from the viewpoint of generalized linear models, which can deal with binary, count, and multiclass data as a response variable.

\section{SVD-based sparse principal component regression}
\label{sec:SPCRsvd}

SPCR consists of basic two loss functions: the squared regression loss function and the PCA loss function by Zou \textit{et al.} (2006). 
However, it is unclear whether the PCA loss is the best for SPCR or not. 
To investigate the issue, we propose another formulation for SPCR by using the SVD approach by Shen and Huang (2008).

We consider the following minimization problem
\begin{align}
& \min_{\beta_0,{\bm \beta}, Z, V} \left\{ \frac{1}{n} \| {\bm y} - \beta_0 {\bm 1}_n - X V {\bm \beta} \|_2^2 + \frac{w}{n} \| X - Z V^T \|_F^2 + \lambda_V \| V \|_1 + \lambda_{\bm \beta} \| \bm \beta \|_1 \right\} \nonumber \\
& {\rm subject \ to} \quad V^T V = I_k,
\label{eq:eq1}
\end{align}
where $\beta_0$ is an intercept, $k$ is the number of PCs, ${\bm \beta}$ is a $k$-dimensional coefficient vector, $Z$ is an $n \times k$ matrix of PCs, $V$ is a $p \times k$ PC loading matrix, and $\bm 1_n$ is an $n$-dimensional vector of which all elements are one. 
In addition, $w \ (\geq 0)$ is a tuning parameter and $\lambda_V,\lambda_{\bm \beta}$ are regularization parameters with non-negative values.

The first term is the least squared loss function between the response and the PCs $XV$. 
The second term is the PCA loss function in the SVD approach by Shen and Huang (2008). 
Although the formula is seemingly different from the first term in Formula \eqref{eq:spcaSVD_first}, these are essentially equivalent: our approach aims to estimate the $k$ PCs simultaneously, while Shen and Huang (2008) estimate sequentially. 
The third and fourth terms are the lasso penalty that induces zero estimates of the parameters $V$ and $\bm \beta$, respectively. 
The tuning parameter $w$ controls the degree of the second term. 
A smaller value for $w$ is used when we aim to obtain better prediction accuracies, while a larger value for $w$ is used when we aim to obtain the exact expression
of the PC loadings. 
The minimization problem \eqref{eq:eq1} enables us to perform regression analysis and PCA simultaneously. 
We call this procedure SPCRsvd.
In Section 5, we will confirm that SPCRsvd is competitive with or better than SPCR through numerical studies.

We remark two points here. 
First, it is possible to use $Z$ in the first term of \eqref{eq:eq1} instead of $XV$, since $Z$ is also the PCs. 
However, the formulation by $Z$ instead of $XV$ did not perform well in numerical studies. 
We, then, adopt the formulation by $XV$. 
Second, SPCR imposes the ridge penalty for the PC loading, but SPCRsvd does not. 
The ridge penalty is basically from SPCA by Zou \textit{et al.} (2006). 
Because SPCRsvd is not based on SPCA by Zou \textit{et al.} (2006), we do not add the ridge penalty in Formula \eqref{eq:eq1}. 
It is possible to add the ridge penalty and replace the lasso penalty with other penalties that induce sparsity, e.g., the adaptive lasso penalty, the SCAD penalty, and minimax concave penalty (Zhang, 2010), but our aim of this paper is to establish the basic procedure of Formula \eqref{eq:eq1}.

%%%%%%%%%%%%%%%%%%%%%%%%%%%%%%%%%%%%%%%%%%%%%%%%%%%%%%%%%%%%%%%%%%%%%%%%%%%%%%%%%%%%%%%%%%%%%%%%%%%%%%%%%%%%%%%%%%%%%%%%%%%%%%%%%%%%%%%%%%%%%%%%%%%%%%%%%%%%%%%%
\section{Implementation}
\label{sec:implementation}

%%%%%%%%%%%%%%%%%%%%%%%%%%%%%%%%%%%%%%%%%%%%%%%%%%%%%%%%%%%%%%%%%%%%%%%%%%%%%%%%%%
\subsection{Computational algorithm}
\label{sec:algorithm}

To obtain the estimates of the parameters ${\bm \beta}, Z, V$ in Formula \eqref{eq:eq1}, we employ an alternating direction method of multipliers (ADMM) and a linearized alternating direction method of multipliers (LADMM). 
ADMM and LADMM are used in various models with sparse regularization: for example, see Boyd \textit{et al.} (2011), Ye and Xie (2011), Danaher \textit{et al.} (2014), Li \textit{et al.} (2014), Tan \textit{et al.} (2014), Ma and Huang (2017), Yan and Bien (2018), Wang \textit{et al.} (2018), and Price \textit{et al.} (2019). 
To solve the minimization problem \eqref{eq:eq1} by using ADMM, we rewrite the problem as
\begin{align}
& \min_{\beta_0, {\bm \beta}, {\bm \beta}_0, Z, V, V_0, V_1} \left\{ \frac{1}{n} \| {\bm y} - \beta_0 {\bm 1}_n - X V_1 {\bm \beta} \|_2^2 + \frac{w}{n} \| X - Z V^T \|_F^2 + \lambda_V \| V_0 \|_1 + \lambda_{\bm \beta} \| \bm \beta_0 \|_1 \right\} \nonumber \\
& {\rm subject \ to} \quad V^T V = I_k, \quad V=V_0=V_1, \quad \bm \beta = \bm \beta_0.
\label{eq:eq1_AddConstrant1}
\end{align}
The scaled augmented Lagrangian for the problem (\ref{eq:eq1_AddConstrant1}) is then given by
\begin{align*}
&\frac{1}{n} \| {\bm y} - \beta_0 {\bm 1}_n - X V_1 {\bm \beta} \|_2^2 + \frac{w}{n} \| X - Z V^T \|_F^2 + \lambda_V \| V_0 \|_1 + \lambda_{\bm \beta} \| \bm \beta_0 \|_1 \\
& + \frac{\rho_1}{2} \| V - V_0  + \Lambda_1 \|_F^2 +\frac{\rho_2}{2} \| V_1 - V_0 + \Lambda_2 \|_F^2 + \frac{\rho_3}{2} \| \bm \beta - {\bm \beta}_0 + {\bm \lambda}_3 \|_2^2 \\
&{\rm subject \ to} \quad V^T V = I_k,
\end{align*}
where $\Lambda_1, \Lambda_2, {\bm \lambda}_3$ are dual variables and $\rho_1, \rho_2, \rho_3 \ (>0)$ are penalty parameters. 
This leads to the ADMM algorithm as follows:
\begin{description}
%%%
\item[Step 1] Set the values of the tuning parameter $w$, the regularization parameters $\lambda_V, \lambda_{\bm \beta}$, and the penalty parameters $\rho_1, \rho_2, \rho_3$. 
%%%
\item[Step 2] Initialize the all parameters by $\beta_0^{(0)}, {\bm \beta}^{(0)}, {\bm \beta}_0^{(0)}, Z^{(0)}, V^{(0)}, V_0^{(0)}, V_1^{(0)},\Lambda_1^{(0)}, \Lambda_2^{(0)}, {\bm \lambda}_3^{(0)}$. 
%%%
\item[Step 3] For $m=0,1,2,\ldots$, repeat from Step 4 to Step 11 until convergence.
%%%
\item[Step 4] Update $V_1$ as follows:
\begin{align*}
{\rm vec} (V_1^{(m+1)}) &= \left( \frac{1}{n} {\bm \beta}^{(m)} {\bm \beta}^{(m)T} \otimes X^T X + \frac{\rho_2}{2} I_k \otimes I_p \right)^{-1} {\rm vec} \bigg\{ \frac{1}{n} X^T ({\bm y} - \beta_0^{(m)} {\bm 1}_n) {\bm \beta}^{(m)T} \\
& \hspace{5mm}  + \frac{\rho_2}{2} (V_0^{(m)} - \Lambda_2^{(m)}) \bigg\},
\end{align*}
where $\otimes$ represents the Kronecker product. 
%%%
\item[Step 5] Update $V$ as follows:
\begin{equation*}
V^{(m+1)}=PQ^T,
\end{equation*}
where $P$ and $Q$ are the matrices given by the SVD 
\begin{equation*}
\displaystyle{\frac{w}{n} X^T Z^{(m)} + \frac{\rho_1}{2} \left(V_0^{(m)} - \Lambda_1^{(m)} \right) = P \Omega Q^T}.
\end{equation*}
%%%
\item[Step 6] Update $V_0$ as follows:
\begin{equation*}
v_{0ij}^{(m+1)} = {\mathcal S} \left( \frac{\rho_1(v_{ij}^{(m+1)} + \lambda_{1ij}^{(m)}) +\rho_2(v_{ij}^{(m+1)} + \lambda_{2ij}^{(m)})}{\rho_1 + \rho_2}, \frac{\lambda_V}{\rho_1 + \rho_2} \right), \quad i=1,\ldots,p, \ j=1,\ldots,k,
%\label{eqn:V0}
\end{equation*}
where $v_{0ij}^{(m)}=(V_0^{(m)})_{ij}$, $v_{ij}^{(m)}=(V^{(m)})_{ij}$, $\lambda_{\ell ij} \ (\ell =1,2)$ is the $(i,j)$-th element of the matrix $\Lambda_\ell \ (\ell=1,2)$, and ${\mathcal S} (\cdot,\cdot) $ is the soft-thresholding operator defined by ${\mathcal S}(x,\lambda)={\rm sign}(x)(|x|-\lambda)_+$.
%%%
\item[Step 7] Update $Z$ by $Z^{(m+1)}=X V^{(m+1)}$.
%%%
\item[Step 8] Update ${\bm \beta}$ as follows:
\begin{equation*}
\bm \beta^{(m+1)} = \left(\frac{1}{n} V_1^{(m+1)T} X^T X V_1^{(m+1)} + \frac{\rho_3}{2} I_k \right)^{-1} \left\{ \frac{1}{n} V_1^{(m+1)T} X^T ({\bm y} - \beta_0^{(m)} {\bm 1}_n) + \frac{\rho_3}{2} ({\bm \beta}_0^{(m)} - {\bm \lambda}^{(m)}_3)  \right\}.
\end{equation*}
%%%
\item[Step 9] Update ${\bm \beta}_0$ as follows:
\begin{equation*}
\beta_{0j}^{(m+1)} = {\mathcal S} \left( \beta_j^{(m+1)} + \lambda_{3j}^{(m)}, \frac{\lambda_\beta}{\rho_3} \right), \quad j=1,\ldots,k,
%\label{eqn:beta0}
\end{equation*}
where $\lambda_{3j}^{(m)}$ and $\beta_j^{(m)}$ are the $j$-th element of the vector ${\bm \lambda}_3^{(m)}$ and ${\bm \beta}^{(m)}$, respectively. 
%%%
\item[Step 10] Update $\beta_0$ as follows:
\begin{equation*}
\beta_{0}^{(m+1)} = \frac{1}{n} {\bm 1}_n^T ( {\bm y} - X V_1^{(m+1)} {\bm \beta}^{(m+1)} ).
%\label{eqn:beta0}
\end{equation*}
%%%
\item[Step 11] Update ${\Lambda}_1,{\Lambda}_2,{\bm \lambda}_3$ as follows:
\begin{align*}
\Lambda_1^{(m+1)} &= \Lambda_1^{(m)} + V^{(m+1)} - V_0^{(m+1)},\\
\Lambda_2^{(m+1)} &= \Lambda_2^{(m)} + V_1^{(m+1)} - V_0^{(m+1)},\\
{\bm \lambda}_3^{(m+1)} &= {\bm \lambda}_3^{(m)} + {\bm \beta}^{(m+1)} - {\bm \beta}^{(m+1)}_0.
\end{align*}
\end{description}
The derivation of the updates is given in Appendix A.

To apply LADMM into the minimization problem (\ref{eq:eq1}), we consider the following problem
\begin{align}
& \min_{\beta_0, {\bm \beta}, {\bm \beta}_0, Z, V, V_0} \left\{ \frac{1}{n} \| {\bm y} - \beta_0 {\bm 1}_n - X V_0 {\bm \beta} \|_2^2 + \frac{w}{n} \| X - Z V^T \|_F^2 + \lambda_V \| V_0 \|_1 + \lambda_{\bm \beta} \| \bm \beta_0 \|_1 \right\} \nonumber \\
& {\rm subject \ to} \quad V^T V = I_k, \quad V=V_0, \quad \bm \beta = \bm \beta_0.
\label{eq:eq1_AddConstrant2}
\end{align}
The augmented Lagrangian for this problem is then given by
\begin{align*}
&\frac{1}{n} \| {\bm y} - \beta_0 {\bm 1}_n - X V_0 {\bm \beta} \|_2^2 + \frac{w}{n} \| X - Z V^T \|_F^2 + \lambda_V \| V_0 \|_1 + \lambda_{\bm \beta} \| \bm \beta_0 \|_1 \\
& + \frac{\rho_1}{2} \| V_0 - V  + \Lambda \|_F^2  + \frac{\rho_2}{2} \| \bm \beta - {\bm \beta}_0 + {\bm \lambda} \|_2^2 \\
& {\rm subject \ to} \quad V^T V = I_k,
\end{align*}
where $\Lambda, {\bm \lambda}$ are dual variables and $\rho_1, \rho_2 \ (>0)$ are penalty parameters. 
The updates of the LADMM algorithm is almost same with those of the ADMM algorithm. 
We summarize the updates and the derivation in Appendix B.

%%%%%%%%%%%%%%%%%%%%%%%%%%%%%%%%%%%%%%%%%%%%%%%%%%%%%%%%%%%%%%%%%%%%%%%%%%%%%%%%%%
\subsection{Determination of tuning parameters}
\label{sec:DeterminationTuningParameters}
We have the six tuning parameters: $w, \lambda_V, \lambda_{\bm \beta}, \rho_1, \rho_2, \rho_3$. 
The penalty parameters $\rho_1, \rho_2, \rho_3$ are fixed as $\rho_1=\rho_2=\rho_3=1$ according to Boyd \textit{et al.} (2011). 
The tuning parameter $w$ is set according to the purpose of the analysis. 
A small value is allocated to the value for $w$, if a user considers that the regression loss is more important than the PCA loss. 
This idea follows Kawano \textit{et al.} (2015; 2018).

The two regularization parameters $\lambda_V, \lambda_{\bm \beta}$ are objectively selected by $K$-fold cross-validation. 
When we have divided $K$ datasets $({\bm y}^{(1)}, X^{(1)}), \ldots, ({\bm y}^{(K)}, X^{(K)})$ from the original dataset, the criterion for the $K$-fold cross-validation in ADMM is given by
\begin{equation}
{\rm CV} = \frac{1}{K} \sum_{k=1}^K \frac{1}{n} \left\| {\bm y}^{(k)} - \hat{\beta}_0^{(-k)} {\bm 1}_{(k)} - X^{(k)} \hat{V}_1^{(-k)} \hat{\bm \beta}^{(-k)} \right\|_2^2,
\label{eq:CV}
\end{equation}
where $\hat{\beta}_0^{(-k)}, \hat{V}_1^{(-k)}, \hat{\bm \beta}^{(-k)}$ are the estimates of ${\beta}_0, {V}_1, {\bm \beta}$, respectively, computed with the data removing the $k$-th part. 
We omit the CV criterion for LADMM, since we only replace $\hat{V}_1^{(-k)}$ in \eqref{eq:CV} into $\hat{V}_0^{(-k)}$. 
In our numerical studies, we set $K=5$.

%To prepare candidate values of $\lambda_V, \lambda_{\bm \beta}$, we need to find a maximum value of the two regularization parameters respectively. 
%From , the maximum value of $\lambda_V$ is 
%
%
%
%\noindent \textbf{Determination of a maximum value of $\lambda_{V}$: $\lambda_{V, {\rm max}}$}
%
%From (\ref{eqn:V0}), we obtain
%\begin{equation*}
%\lambda_{V, {\rm max}} = ( \rho_1 + \rho_2 ) \max_{i,j} | v_{ij}^{1} |,
%\end{equation*}
%where $v_{ij}^{1}$ is an $(i,j)$-th element of a matrix $V^1$. 
%$V^1$ is given by $V^1 = PQ^T$, where $\displaystyle{\frac{w}{n} X^T Z^0 + \frac{\rho_1}{2} V_0^0 = P \Lambda Q^T}$, and $V_0^0$ and $Z^0$ are the initial values of $V_0$ and $Z$ respectively. 
%
%\hspace{2mm}
%
%\noindent \textbf{Determination of a maximum value of $\lambda_{\beta}$: $\lambda_{\beta, {\rm max}}$}
%
%From (\ref{eqn:beta0}), we obtain
%\begin{equation*}
%\lambda_{\beta, {\rm max}} =  \rho_3  \max_{j} | \beta_{j}^{1} |,
%\end{equation*}
%where $\beta_{j}^{1}$ is a $j$-th element of a vector ${\bm \beta}^1$. 
%${\bm \beta}^1$ is given by
%\begin{equation*}
%{\bm \beta}^1 =  \left(\frac{1}{n} (V_1^0)^T X^T X V_1^0 + \frac{\rho_3}{2} I_k \right)^{-1} \frac{1}{n} (V_1^0)^T X^T {\bm y},
%\end{equation*}
%where $V_1^0$ is the initial value of $V_1$. 
%This determination is according to that of CDA. 

%%%%%%%%%%%%%%%%%%%%%%%%%%%%%%%%%%%%%%%%%%%%%%%%%%%%%%%%%%%%%%%%%%%%%%%%%%%%%%%%%%%%%%%%%%%%%%%%%%%%%%%%%%%%%%%%%%%%%%%%%%%%%%%%%%%%%%%%%%%%%%%%%%%%%%%%%%%%%%%%
\section{Numerical study}
\label{sec:NumericalStudy}

%%%%%%%%%%%%%%%%%%%%%%%%%%%%%%%%%%%%%%%%%%%%%%%%%%%%%%%%%%%%%%%%%%%%%%%%%%%%%%%%%%
\subsection{Monte Carlo simulations}
\label{sec:MonteCarlo}

We conducted Monte Carlo simulations to investigate the effectiveness of SPCRsvd. 
The simulations have five cases, which are the same as Kawano \textit{et al.} (2015). 
The five cases are given as follows.
\begin{description}
%%%%%%%%%%%%%%
\item[Case 1:]  The 10-dimensional covariate vector ${\bm x}=(x_1,\ldots,x_{10})$ was according to a multivariate normal distribution having mean zero vector and variance-covariance matrix $\Sigma$. 
The response was obtained by
\begin{equation}
y_i = \zeta_1 x_{i1} + \zeta_2 x_{i2} + \varepsilon_i, \quad i=1,\ldots,n,
\label{eq:case1_MonteCarlo}
\end{equation}
where $\varepsilon_i$ is independently distributed as  a normal distribution having mean zero and variance $\sigma^2$. 
We used $\zeta_1=2, \zeta_2=1, \Sigma=I_{10}$.  
%%%%%%%%%%%%%%
\item[Case 2:]  This case is the same as Case 1 except for $\zeta_1=8, \zeta_2=1, \Sigma={\rm diag} (1,3^2,1,\ldots,1)$. 
%%%%%%%%%%%%%%
\item[Case 3:]  The 20-dimensional covariate vector ${\bm x}=(x_1,\ldots,x_{20})$ was according to a multivariate normal distribution $N_{20}({\bm 0}, \Sigma)$. 
The response was obtained by
\begin{equation}
y_i = 4 {\bm \zeta}^T {\bm x}_i + \varepsilon_i, \quad i=1,\ldots,n,
\label{eq:case1_MonteCarlo}
\end{equation}
where $\varepsilon_i$ is independently distributed as $N(0,\sigma^2)$. 
We used ${\bm \zeta}=({\bm \nu}, 0,\ldots,0)^T$ and $\Sigma={\rm block \ diag}(\Sigma_1, I_{11})$, where ${\bm \nu}=(-1,0,1,1,0,-1,-1,0,1)$ and $\left(\Sigma_1 \right)_{ij}=0.9^{|i-j|} \ (i,j,=1,\ldots,9)$. 
%%%%%%%%%%%%%%
\item[Case 4:] The 30-dimensional covariate vector ${\bm x}=(x_1,\ldots,x_{30})$ was according to a multivariate normal distribution $N_{30}({\bm 0}, \Sigma)$. 
The response was obtained by
\begin{equation}
y_i = 4 {\bm \zeta}_1^T {\bm x}_i + 4 {\bm \zeta}_2^T {\bm x}_i + \varepsilon_i, \quad i=1,\ldots,n,
\label{eq:case1_MonteCarlo}
\end{equation}
where $\varepsilon_i$ is independently distributed as $N(0,\sigma^2)$. 
We used ${\bm \zeta}_1=({\bm \nu}_1, 0,\ldots,0)^T, {\bm \zeta}_2=(\underbrace{0,\ldots,0}_{9},{\bm \nu}_2, \underbrace{0,\ldots,0}_{15})^T, \Sigma={\rm block \ diag}(\Sigma_1, \Sigma_2, I_{15})$. 
Here ${\bm \nu}_1=(-1,0,1,1,0,-1,-1,0,1)$, ${\bm \nu}_2=(\underbrace{1,\ldots,1}_{6})$, and $\left(\Sigma_\ell \right)_{ij}=0.9^{|i-j|} \ (i,j,=1,\ldots,9; \ell=1,2)$. 
%%%%%%%%%%%%%%
\item[Case 5:] This case is the same as Case 4 except for $\bm \nu_2=(1,0,-1,-1,0,1)$.
\end{description}
The details of the setteings are referred to Kawano \textit{et al.} (2015).

The sample size was set to $n=50, 200$. 
The standard deviation was set to $\sigma=1,2$. 
SPCRsvd was fitted to the simulated data with one or five components $(k=1,5)$. 
We set the value of the tuning parameter $w$ to 0.1.  
We considered two algorithms in Section \ref{sec:algorithm}: ADMM for SPCRsvd (SPCRsvd-ADMM) and LADMM for SPCRsvd (SPCRsvd-LADMM). 
SPCRsvd was compared with SPCR, PCR, sparse partial least squares (SPLS) by Chun and Kele\c{s} (2010), and partial least squares (PLS) by Wold (1975). 
SPCR was computed by the package \textbf{spcr}, SPLS by \textbf{spls}, and PLS and PCR by \textbf{pls}. 
These packages are included in \textsf{R} (R Core Team, 2020). 
The values of the tuning parameters $w$ and $\xi$ in SPCR were set to 0.1 and 0.01, respectively, and then the regularization parameters were selected by five-fold cross-validation. 
The values of tuning parameters in SPLS, PLS, and PCR were selected by 10-fold cross-validation. 
The performance was evaluated in terms of ${\rm MSE}=E[ (y-\hat{y})^2 ]$. 
The simulation was conducted 100 times. 
MSE was estimated by 1,000 random samples.

We summarize the means and the standard deviations of MSEs from Table \ref{table:case1} to Table \ref{table:case5}. 
The results for $\sigma=1,2$ had similar tendencies. 
PCR and PLS were worst in almost all cases. 
First, we discuss the results among SPCRsvd-LADMM, SPCRsvd-ADMM, and SPCR. 
SPCRsvd-LADMM and SPCRsvd-ADMM were competitive with SPCR. 
In particular, SPCRsvd-LADMM and SPCRsvd-ADMM provided smaller MSEs than SPCR in almost all cases when $k=1$. 
Note that SPCR provided large values of standard deviation in some cases. 
This means that SPCR sometimes produces so large value of MSE. 
This fact can cause instability of SPCR. 
Compared to SPLS, SPCRsvd-LADMM and SPCRsvd-ADMM were slightly inferior in many cases when $k=5$. 
However, SPLS produced so large values of MSEs in many cases when $k=1$.  
From this experiment, we observed that SPCRsvd-LADMM and SPCRsvd-ADMM provided relatively stable smaller values of MSEs than other methods. %, while our proposed methods 

%\begin{landscape}
\begin{table}[htbp]
\begin{center}
\small
\caption{Mean (standard deviation) values of the MSE for Case 1.
The bold values correspond to the smallest means among SPCRsvd-LADMM, SPCRsvd-ADMM, and SPCR.  }
%\vspace{5mm}
%\scalebox{0.9}[0.9]{
%\begin{tabular}{@{\extracolsep{-5.5pt}}ccccccccc} \hline
\begin{tabular}{ccccccccc} \hline
$\sigma$ & $n$ & $k$ & SPCRsvd-LADMM & SPCRsvd-ADMM & SPCR & SPLS & PLS & PCR  \\
\hline
1 & 50 & 1 & 1.302 & \textbf{1.192} & 1.814 & 1.507 & 2.062 & 5.735 \\
 &  &  & (0.722) & (0.196) & (1.596) & (0.476) & (0.514) & (0.598) \\
 &  & 5 & 1.309 & \textbf{1.227} & 1.377 & {1.168} & 1.313 & 3.721 \\
 &  &  & (0.727) & (0.215) & (0.849) & (0.217) & (0.173) & (1.118) \\
 \hline
 & 200 & 1 & 1.235 & \textbf{1.027} & 3.738 & {1.023} & 1.242 & 5.511\\
 &  &  & (0.979) & (0.055) & (2.472) & (0.055) & (0.1208) & (0.683))\\
 &  & 5 & 1.144 & \textbf{1.042} & 2.310 & {1.015} & 1.050 & 3.486\\
 &  &  & (0.717) & (0.058) & (2.176) & (0.051) & (0.049) & (1.075)\\
 \hline
2 & 50 & 1 & 5.227 & \textbf{4.875} & 5.648 & 5.104 & 5.522 & 8.834 \\
 &  &  & (1.322) & (0.604) & (1.608) & (0.806) & (0.722) & (0.716) \\
 &  & 5 & \textbf{4.892} & 4.902 & 5.109 & {4.840} & 5.259 & 7.054 \\
 &  &  & (0.743) & (0.605) & (0.852) & (0.827) & (0.717) & (1.201) \\
 \hline
 & 200 & 1 & 4.490 & \textbf{4.113} & 6.959 & {4.052} & 4.355 & 8.501 \\
 &  &  & (1.276) & (0.200) & (2.381) & (0.211) & (0.228) & (0.741) \\
 &  & 5 & 4.231 & \textbf{4.148} & 5.343 & {4.070} & 4.201 & 6.533 \\
 &  &  & (0.673) & (0.198) & (2.121) & (0.205) & (0.199) & (1.099) \\
\hline
\end{tabular}
\label{table:case1}
\end{center}
%}
\end{table}
%\end{landscape}

%\begin{landscape}
\begin{table}[htbp]
\begin{center}
\small
\caption{Mean (standard deviation) values of the MSE for Case 2.
The bold values correspond to the smallest means among SPCRsvd-LADMM, SPCRsvd-ADMM, and SPCR.  }
%\vspace{5mm}
%\scalebox{0.9}[0.9]{
%\begin{tabular}{@{\extracolsep{-5.5pt}}ccccccccc} \hline
\begin{tabular}{ccccccccc} \hline
$\sigma$ & $n$ & $k$ & SPCRsvd-LADMM & SPCRsvd-ADMM & SPCR & SPLS & PLS & PCR  \\
\hline
1 & 50 & 1 & 1.230 & \textbf{1.226} & 1.318 & 40.881 & 47.496 & 67.451 \\
 &  &  & (0.195) & (0.152) & (0.179) & (19.390) & (10.342) & (3.768) \\
 &  & 5 & 1.284 & 1.245 & \textbf{1.188} & {1.130} & 1.346 & 39.351 \\
 &  &  & (0.260) & (0.160) & (0.173) & (0.162) & (0.214) & (15.154) \\
 \hline
 & 200 & 1 & \textbf{1.032} & 1.036 & 1.050 & 43.161 & 47.591 & 65.540 \\
 &  &  & (0.059) & (0.055) & (0.050) & (13.773) & (4.135) & (2.947) \\
 &  & 5 & 1.058 & 1.038 & \textbf{1.022} & {1.017} & 1.050 & 35.790 \\
 &  &  & (0.080) & (0.062) & (0.048) & (0.051) & (0.049) & (12.650) \\
 \hline
2 & 50 & 1 & 5.709 & \textbf{5.019} & 5.948 & 43.762 & 50.631 & 70.595 \\
 &  &  & (6.809) & (0.630) & (6.860) & (19.637) & (10.344) & (4.001) \\
 &  & 5 & 5.281 & 5.128 & \textbf{4.803} & {4.558} & 5.291 & 42.679 \\
 &  &  & (0.921) & (0.706) & (0.654) & (0.740) & (0.757) & (15.099) \\
 \hline
 & 200 & 1 & \textbf{4.161} & 4.174 & 4.199 & 46.133 & 50.468 & 68.517 \\
 &  &  & (0.230) & (0.216) & (0.199) & (13.786) & (4.305) & (3.071) \\
 &  & 5 & 4.207 & 4.180 & \textbf{4.088} & {4.065} & 4.201 & 38.810 \\
 &  &  & (0.269) & (0.231) & (0.192) & (0.203) & (0.199) & (12.594) \\
\hline
\end{tabular}
\label{table:case2}
\end{center}
%}
\end{table}
%\end{landscape}

%\begin{landscape}
\begin{table}[htbp]
\begin{center}
\small
\caption{Mean (standard deviation) values of the MSE for Case 3.
The bold values correspond to the smallest means among SPCRsvd-LADMM, SPCRsvd-ADMM, and SPCR.  }
%\vspace{5mm}
%\scalebox{0.9}[0.9]{
%\begin{tabular}{@{\extracolsep{-5.5pt}}ccccccccc} \hline
\begin{tabular}{ccccccccc} \hline
$\sigma$ & $n$ & $k$ & SPCRsvd-LADMM & SPCRsvd-ADMM & SPCR & SPLS & PLS & PCR  \\
\hline
1 & 50 & 1 & \textbf{1.564} & 1.581 & 1.793 & 20.625 & 20.847 & 21.404 \\
 &  &  & (0.314) & (0.331) & (2.160) & (1.924) & (2.012) & (1.295) \\
 &  & 5 & 1.663 & 1.933 & \textbf{1.563} & 1.998 & 3.398 & 22.244 \\
 &  &  & (0.437) & (0.602) & (0.316) & (1.192) & (1.442) & (1.475) \\
 \hline
 & 200 & 1 & \textbf{1.085} & 1.098 & 1.096 & 15.259 & 16.817 & 20.642 \\
 &  &  & (0.068) & (0.072) & (0.069) & (4.717) & (2.886) & (0.863) \\
 &  & 5 & 1.114 & 1.144 & \textbf{1.096} & {1.089} & 1.158 & 20.759 \\
 &  &  & (0.083) & (0.105) & (0.070) & (0.240) & (0.080) & (0.917) \\
 \hline
2 & 50 & 1 & 6.412 & \textbf{6.408} & 6.562 & 24.353 & 24.423 & 24.520 \\
 &  &  & (1.279) & (1.247) & (2.057) & (2.389) & (2.342) & (1.441) \\
 &  & 5 & 6.615 & 6.829 & \textbf{6.349} & 6.525 & 8.000 & 25.519 \\
 &  &  & (1.591) & (1.832) & (1.258) & (2.178) & (2.183)& (1.730) \\
 \hline
 & 200 & 1 & \textbf{4.579} & 4.610 & 4.766 & 19.078 & 20.220 & 23.627 \\
 &  &  & (1.963) & (1.961) & (2.632) & (4.390) & (2.733) & (1.002) \\
 &  & 5 & 4.654 & \textbf{4.451} & 4.763 & {4.272} & 4.430 & 23.776 \\
 &  &  & (1.963) & (0.300) & (2.632) & (0.361) & (0.272) & (1.063) \\
\hline
\end{tabular}
\label{table:case3}
\end{center}
%}
\end{table}
%\end{landscape}

%\begin{landscape}
\begin{table}[htbp]
\begin{center}
\small
\caption{Mean (standard deviation) values of the MSE for Case 4.
The bold values correspond to the smallest means among SPCRsvd-LADMM, SPCRsvd-ADMM, and SPCR.  }
%\vspace{5mm}
%\scalebox{0.9}[0.9]{
%\begin{tabular}{@{\extracolsep{-5.5pt}}ccccccccc} \hline
\begin{tabular}{ccccccccc} \hline
$\sigma$ & $n$ & $k$ & SPCRsvd-LADMM & SPCRsvd-ADMM & SPCR & SPLS & PLS & PCR  \\
\hline
1 & 50 & 1 & 2.595 & \textbf{2.302} & 2.307 & 21.540 & 47.460 & 433.826 \\
 &  &  & (1.542) & (0.593) & (0.619) & (1.389) & (23.355) & (114.041) \\
 &  & 5 & 2.720 & 2.646 & \textbf{2.249} & 6.0157 & 11.939 & 33.604 \\
 &  &  & (1.903) & (0.819) & (0.558) & (5.308) & (3.919) & (7.875) \\
 \hline
 & 200 & 1 & 1.160 & 1.176 & \textbf{1.158} & 21.018 & 24.899 & 477.828 \\
 &  &  & (0.075) & (0.077) & (0.076) & (0.991) & (5.165) & (37.972) \\
 &  & 5 & 1.165 & 1.201 & \textbf{1.158} & 1.183 & 1.701 & 23.414 \\
 &  &  & (0.079) & (0.103) & (0.077) & (0.106) & (0.261) & (1.582) \\
 \hline
2 & 50 & 1 & 9.695 & \textbf{9.511} & 9.667 & 24.983 & 50.747 & 437.040 \\
 &  &  & (3.176) & (2.290) & (2.413) & (1.946) & (23.481) & (114.199) \\
 &  & 5 & 10.734 & 9.552 & \textbf{9.511} & 12.712 & 17.237 & 36.904 \\
 &  &  & (4.010) & (2.480) & (2.320) & (6.581) & (4.258) & (7.903) \\
 \hline
 & 200 & 1 & 4.705 & 4.695 & \textbf{4.662} & 24.103 & 27.978 & 480.882 \\
 &  &  & (0.304) & (0.303) & (0.310) & (1.213) & (5.196) & (37.853) \\
 &  & 5 & 4.764 & 4.744 & \textbf{4.660} & {4.458} & 5.219 & 26.522 \\
 &  &  & (0.390) & (0.319) & (0.312) & (0.305) & (0.462) & (1.730) \\
\hline
\end{tabular}
\label{table:case4}
\end{center}
%}
\end{table}
%\end{landscape}

%\begin{landscape}
\begin{table}[htbp]
\begin{center}
\small
\caption{Mean (standard deviation) values of the MSE for Case 5.
The bold values correspond to the smallest means among SPCRsvd-LADMM, SPCRsvd-ADMM, and SPCR.  }
%\vspace{5mm}
%\scalebox{0.9}[0.9]{
%\begin{tabular}{@{\extracolsep{-5.5pt}}ccccccccc} \hline
\begin{tabular}{ccccccccc} \hline
$\sigma$ & $n$ & $k$ & SPCRsvd-LADMM & SPCRsvd-ADMM & SPCR & SPLS & PLS & PCR  \\
\hline
1 & 50 & 1 & 2.155 & 2.207 & \textbf{2.144} & 35.283 & 35.094 & 34.654 \\
 &  &  & (0.501) & (0.577) & (0.524) & (3.264) & (2.726) & (1.806) \\
 &  & 5 & 2.574 & 3.171 & \textbf{2.113} & 10.190 & 16.033 & 35.537 \\
 &  &  & (1.142) & (1.654) & (0.512) & (6.852) & (5.439) & (2.125) \\
 \hline
 & 200 & 1 & \textbf{1.151} & 1.493 & 1.506 & 30.629 & 30.876 & 34.297 \\
 &  &  & (0.076) & (3.318) & (3.491) & (2.614) & (2.497) & (1.602) \\
 &  & 5 & 1.208 & 1.220 & \textbf{1.156} & 1.236 & 1.814 & 34.208 \\
 &  &  & (0.112) & (0.105) & (0.076) & (0.167) & (0.304) & (1.580) \\
 \hline
2 & 50 & 1 & \textbf{8.949} & 8.985 & 9.236 & 38.659 & 38.612 & 37.800 \\
 &  &  & (2.151) & (2.133) & (3.420) & (3.369) & (3.104) & (2.004) \\
 &  & 5 & 9.671 & 9.405 & \textbf{8.848} & 17.133 & 21.495 & 38.780 \\
 &  &  & (2.993) & (2.447) & (2.170) & (8.875) & (6.056) & (2.355) \\
 \hline
 & 200 & 1 & \textbf{4.654} & 4.675 & 4.999 & 34.093 & 34.301 & 37.374 \\
 &  &  & (0.300) & (0.306) & (3.575) & (2.778) & (2.681) & (1.812) \\
 &  & 5 & 4.805 & 4.719 & \textbf{4.635} & {4.555} & 5.306 & 37.343 \\
 &  &  & (0.376) & (0.322) & (0.307) & (0.458) & (0.471) & (1.790) \\
\hline
\end{tabular}
\label{table:case5}
\end{center}
%}
\end{table}
%\end{landscape}

\begin{table}[htbp]
\begin{center}
\small
\caption{Mean (standard deviation) values of TPR and TNR for Case 1.
The bold values correspond to the largest means. }
%\vspace{5mm}
%\scalebox{0.9}[0.9]{
\renewcommand{\arraystretch}{0.85}
%\begin{tabular}{@{\extracolsep{-5.5pt}}cccccccc} \hline
\begin{tabular}{cccccccc} \hline
$\sigma$ & $n$ & $k$ &  & SPCRsvd-LADMM & SPCRsvd-ADMM & SPCR & SPLS \\
\hline
1 & 50 & 1 & TPR & 0.980 & \textbf{1} & 0.880 & 0.870 \\
 &  &  &  & (0.140) & (0) & (0.326) & (0.220) \\
 &  &  & TNR & 0.578 & 0.657 & 0.510 & \textbf{0.979} \\
 &  &  &  & (0.218) & (0.190) & (0.212) & (0.056) \\
 &  & 5 & TPR & 0.980 & \textbf{1} & 0.970 & 0.995 \\
 &  &  &  & (0.140) & 0 & (0.171) & (0.050) \\
 &  &  & TNR & 0.619 & 0.617 & 0.479 & \textbf{0.931} \\
 &  &  &  & (0.207) & (0.195) & (0.150) & (0.127) \\
 \hline
 & 200 & 1 & TPR & 0.960 & \textbf{1} & 0.450 & \textbf{1} \\
 &  &  &  & (0.196) & (0) & (0.500) & (0) \\
 &  &  & TNR & 0.584 & 0.773 & 0.787 & \textbf{1} \\
 &  &  &  & (0.269) & (0.246) & (0.259) & (0) \\
 &  & 5 & TPR & 0.980 & \textbf{1} & 0.740 & \textbf{1} \\
 &  &  &  & (0.140) & (0) & (0.440) & (0) \\
 &  &  & TNR & 0.626 & 0.723 & 0.672 & \textbf{0.959} \\
 &  &  &  & (0.257) & (0.241) & (0.237) & (0.086) \\
 \hline
2 & 50 & 1 & TPR & 0.920 & \textbf{0.990} & 0.850 & 0.790 \\
 &  &  &  & (0.263) & (0.070) & (0.358) & (0.248) \\
 &  &  & TNR & 0.508 & 0.515 & 0.455 & \textbf{0.930} \\
 &  &  &  & (0.200) & (0.168) & (0.242) & (0.133) \\
 &  & 5 & TPR & \textbf{0.980} & 0.955 & \textbf{0.980} & 0.900 \\
 &  &  &  & (0.121) & (0.143) & (0.140) & (0.201) \\
 &  &  & TNR & 0.530 & 0.585 & 0.382 & \textbf{0.910} \\
 &  &  &  & (0.184) & (0.187) & (0.119) & (0.135) \\
 \hline
 & 200 & 1 & TPR & 0.930 & \textbf{1} & 0.410 & \textbf{1} \\
 &  &  &  & (0.256) & (0) & (0.494) & (0) \\
 &  &  & TNR & 0.536 & 0.645 & 0.760 & \textbf{0.994} \\
 &  &  &  & (0.241) & (0.215) & (0.295) & (0.021) \\
 &  & 5 & TPR & 0.980 & \textbf{1} & 0.750 & \textbf{1} \\
 &  &  &  & (0.140) & (0) & (0.435) & (0) \\
 &  &  & TNR & 0.620 & 0.642 & 0.584 & \textbf{0.947} \\
 &  &  &  & (0.215) & (0.216) & (0.262) & (0.107) \\
\hline
\end{tabular}
\renewcommand{\arraystretch}{0.85}
\label{table:case1_TPRTNR}
\end{center}
%}
\end{table}

The true positive rate (TPR) and the true negative rate (TNR) were also computed for SPCRsvd-LADMM, SPCRsvd-ADMM, SPCR, and SPLS.  
TPR and TNR are, respectively, defined by
%where TPR is defined by the ratio that the true coefficients with non-zero values are correctly estimated as non-zero values, while TNR is defined by the ratio that the true coefficients with zero values are correctly estimated as zero. 
\begin{equation*}
\mathrm{TPR}=
\frac{1}{100}
\sum_{k=1}^{100}
\frac{\left|\left\{ 
j:\hat{\zeta}^{(k)}_{j}\neq0~\wedge~\zeta^{\ast}_{j}\neq 0
\right\}\right|}
{\left|\left\{
j:\zeta^{\ast}_{j}\neq 0 
\right\}\right|}, \quad 
\mathrm{TNR}=
\frac{1}{100}
\sum_{k=1}^{100}
\frac{\left|\left\{
j:\hat{\zeta}^{(k)}_{j}=0~\wedge~\zeta^{\ast}_{j}=0
\right\}\right|}
{\left|\left\{
j:\zeta^{\ast}_{j}= 0 
\right\}\right|}, 
\end{equation*}
where ${\zeta}^{*}_{j}$ is the true $j$-th coefficient, $\hat{\zeta}^{(k)}_{j}$ is the estimated $j$-th coefficient for the $k$-th simulation, and $|\{\ast\}|$ is the number of elements included in a set $\{\ast\}$. 
Table \ref{table:case1_TPRTNR} represents the means and standard deviations of TPR and TNR. 
Many methods provided higher ratios of TPRs, whereas SPCR sometimes did not. 
SPLS provided the highest ratios of TNRs in all situations. 
These tendencies were essentially unchanged among all cases. 
The results from Case 2 to Case 5 are shown in the supplementary material. 

%, and present similar results. 
%In Table \ref{Table_TPRTNR_12}, many methods provided higher ratios of TPRs except for aSPCR-Log(1), while aSPCR-Log(1) provided higher ratios of TNRs. 
%The results for Cases 3 and 4 showed that the TPRs were higher in almost all situations, but the TNRs of aSPCR-Poi(0.1) and SPCR-Poi were too much lower. 

%%%%%%%%%%%%%%%%%%%%%%%%%%%%%%%%%%%%%%%%%%%%%%%%%%%%%%%%%%%%%%%%%%%%%%%%%%%%%%%%%%
\subsection{Real data analyses}
\label{sec:RealData}

\begin{table}[t]
\begin{center}
\caption{Sample size and the numbers of covariates in real datasets.}
%\vspace{5mm}
\begin{tabular}{lcc}
\hline
             &  sample size   &  \# of covariates \\ \hline
housing & 506  & 13  \\ 
communities & 1993 &   101 \\ 
concrete & 1030 & 8 \\ 
diabetes & 442 & 10 \\ 
parkinsons & 5875 & 19 \\ 
triazines & 186 & 36 \\ 
winequality-red & 1599 & 11 \\ 
winequality-white & 4898 & 11 \\ 
%sample size & 506   & 768 & 517 &  1030 & 1993    \\
%\# of covariates   & 13 & 8 &  10 &  8 & 101 \\
\hline
\end{tabular}
\label{table:Datasets}
\end{center}
\end{table}

We applied SPCRsvd into real datasets. 
We used eight real datasets: housing, communities, concrete, diabetes, parkinsons, triazines, winequality-red, and winequality-white, which are available from the UCI database (\url{http://archive.ics.uci.edu/ml/index.html}). 
The sample size and the number of covariates are depicted in Table \ref{table:Datasets}. 
If the sample size was larger than 1,100, we randomly extracted 1,100 observations from the dataset. 
For each dataset, we randomly selected 100 observations as training data and remaining as test data to estimate MSEs. 
We standardized the covariates for each dataset. 
We run two algorithms: SPCRsvd-LADMM and SPCRsvd-ADMM. 
The procedure was repeated 50 times.

We compared SPCRsvd with four methods used in Section \ref{sec:MonteCarlo}. 
The number of principal components was set to $k=1$. 
The value of the tuning parameter $w$ in SPCRsvd was set to 0.01, and then $\lambda_V$ and $\lambda_{\bm \beta}$ were selected by five-fold cross-validation. 
The tuning parameters in other methods were selected in similar manners to in Section \ref{sec:MonteCarlo}.

\begin{landscape}
\begin{table}[htbp]
\begin{center}
\small
\caption{Mean (standard deviation) values of the MSE for real datasets. 
The bold values correspond to the smallest means. }
%\vspace{5mm}
%\scalebox{0.9}[0.9]{
%\renewcommand{\arraystretch}{0.85}
%\begin{tabular}{@{\extracolsep{-5.5pt}}cccccccc} \hline
\begin{tabular}[htbp]{lcccccc} \hline
 & SPCRsvd-LADMM & SPCRsvd-ADMM & SPCR & SPLS & PLS & PCR  \\ \hline
housing   & \textbf{28.51} & 28.64 & 28.85 & 33.26 & 29.16 & 29.23 \\
     & (2.85) & (3.15) & (3.06) & (4.67) & (3.24) & (3.23) \\ \hline
communities   & 3.467$\times 10^{-2}$ & 2.802$\times 10^{-2}$ & 3.465$\times 10^{-2}$ & \textbf{2.500}$\times$$\bm {10^{-2}}$ & 7.368$\times 10^{-2}$ & 6.929$\times 10^{-2}$ \\
     & (0.403$\times 10^{-2}$) & (0.627$\times 10^{-2}$) & (0.220$\times 10^{-2}$) & (0.133$\times 10^{-2}$) & (6.501$\times 10^{-2}$) & (4.623$\times 10^{-2}$) \\ \hline
concrete   & 124.4 & \textbf{123.7} & 124.7 & 142.0 & 125.0 & 125.0 \\
     & (13.7) & (13.8) & (14.3) & (11.1) & (14.5) & (14.4) \\ \hline
diabetes   & \textbf{3221} & 3280 & 3280 & 3429 & 3281 & 3282 \\
     & (140) & (163) & (154) & (286) & (156) & (156) \\ \hline
parkinsons   & \textbf{113.6} & 147.5 & 146.2 & 115.9 & 169.6 & 171.6 \\
     & (19.0) & (52.4) & (58.0) & (6.8) & (81.1) & (79.1) \\ \hline
triazines   & 2.510$\times 10^{-2}$ & 2.516$\times 10^{-2}$ & 2.497$\times 10^{-2}$ & \textbf{2.417$\times \bm {10^{-2}}$} & 2.827$\times 10^{-2}$ & 2.798$\times 10^{-2}$ \\
     & (0.370$\times 10^{-2}$) & (0.379$\times 10^{-2}$) & (0.383$\times 10^{-2}$) & (0.332$\times 10^{-2}$) & (0.546$\times 10^{-2}$) & (0.537$\times 10^{-2}$) \\ \hline
winequality-red   & 5.132$\times 10^{-1}$ & 4.875$\times 10^{-1}$ & 4.927$\times 10^{-1}$ & \textbf{4.841$\times \bm {10^{-1}}$} & 4.947$\times 10^{-1}$ & 4.947$\times 10^{-1}$ \\
     & (0.701$\times 10^{-1}$) & (0.516$\times 10^{-1}$) & (0.480$\times 10^{-1}$) & (0.266$\times 10^{-1}$) & (0.451$\times 10^{-1}$) & (0.460$\times 10^{-1}$) \\ \hline
winequality-white   & 6.820$\times 10^{-1}$ & \textbf{6.811$\times \bm {10^{-1}$}} & 6.906$\times 10^{-1}$ & 7.065$\times 10^{-1}$ & 7.012$\times 10^{-1}$ & 7.010$\times 10^{-1}$ \\
     & (0.478$\times 10^{-1}$) & (0.521$\times 10^{-1}$) & (0.325$\times 10^{-1}$) & (0.362$\times 10^{-1}$) & (0.467$\times 10^{-1}$) & (0.472$\times 10^{-1}$) \\ \hline
\end{tabular}
\label{table:RealDataAnal_k1}
\end{center}
%}
\end{table}
\end{landscape}

Table \ref{table:RealDataAnal_k1} indicates the means and standard deviations of MSEs. 
PLS and PCR were competitive and did not provide the smallest MSEs for all datasets. 
SPCR was slightly better than PLS and PCR. 
SPCRsvd-LADMM and SPCRsvd-ADMM provided smaller MSEs than other methods in many cases. 
Although SPLS provided smaller MSEs than other methods, SPLS had the worst MSEs in some cases. 
From the result, we may conclude that SPCRsvd-LADMM and SPCRsvd-ADMM give smaller and more stable MSEs than other methods, which is consistent with the result in Section \ref{sec:MonteCarlo}.

\section{Conclusions}
\label{sec:Conclusions}

We presented SPCRsvd, a one-stage procedure for PCR with the loss functions that combine a regression loss with a PCA loss from the SVD. 
%Sparse regularization was employed to estimate parameters in the model. 
%We called this method sparse principal component regression based on singular value decomposition approach (SPCRsvd). 
To obtain the estimates of the parameters in SPCRsvd, we developed the computational algorithm by using ADMM and LADMM. 
Our one-stage method was competitive or better than competing approaches. 
Specifically, SPCRsvd produced more stable MSEs than SPCR.

A major limitation of SPCRsvd is the computational cost. 
The limitation causes some problems. 
For example, we observe that SPCRsvd provides relatively low ratios of TPR and TNR from Table \ref{table:case1_TPRTNR}. 
To address the issue, the adaptive lasso would be applied in the regularization term in SPCRsvd. 
However, owing to the computational cost, it may be difficult to perform SPCRsvd with the adaptive lasso, because the adaptive lasso generally requires more computational times than lasso.

SPCRsvd cannot treat binary data for the explanatory variables. 
To perform PCA for binary data, Lee \textit{et al.} (2010) introduced the logistic PCA with sparse regularization. 
It is interesting to extend SPCRsvd in the context of the method by Lee \textit{et al.} (2010). 
We leave them as a future research.

%%%%%%%%%%%%%%%%%%%%%%%%%%%%%%%%%%%%%%%%%%%%%%%%%%%%%%%%%%%%%%%%%%%%%%%%%%%%%%%%%%%%%%%%%%%%%%%%%%%%%%%%%%%%%%%%%%%%%%%%%%%%%%%%%%%%%%%%%%%%%%%%%%%%%%%%%%%%%%%%
\section*{Appendix}
\label{sec:Appendix}

%%%%%%%%%%%%%%%%%%%%%%%%%%%%%%%%%%%%%%%%%%%%%%%%%%%%%%%%%%%%%%%%%%%%%%%%%%%%%%%%%%
\renewcommand{\theequation}{A.\arabic{equation}}
\setcounter{equation}{0}
\subsection*{A \quad Derivation of updates in the ADMM algorithm}
%\label{sec:AppendixA}
By simple calculation, we can easily obtain the solutions for $\beta_0, \Lambda_1, \Lambda_2, {\bm \lambda}_3$. 
Hence, we give the derivation for $V_1, V, V_0, Z, {\bm \beta}, {\bm \beta}_0$. 
Also, we omit the index $m$ for iteration to avoid complications. 
\begin{description}
%%%
\item[Update of $V_1$.]
\begin{equation*}
V_1 := \argmin_{V_1} \left\{ \frac{1}{n} \| {\bm y} - \beta_0 {\bm 1}_n - X V_1 {\bm \beta} \|_2^2 + \frac{\rho_2}{2} \left\| V_1 - V_0 + \Lambda_2 \right\|_F^2 \right\}.
\end{equation*}
Set ${\bm y}^* = {\bm y} - \beta_0 {\bm 1}_n$. 
The terms of the right-hand side are, respectively, calculated as
\begin{align*}
\| {\bm y}^* - X V_1 {\bm \beta} \|_2^2 &=  {\bm y}^{*T} {\bm y}^* - 2 {\rm tr} ({\bm \beta} {\bm y}^{*T} X V_1) + {\bm \beta}^T V_1^T X^T X V_1 {\bm \beta}, \\
\| V_1 - V_0 + \Lambda_2 \|_F^2 &= {\rm tr} (V_1^T V_1) - 2 {\rm tr} \{ (V_0-\Lambda_2)^T V_1\} + {\rm tr}\{ (V_0-\Lambda_2)^T (V_0-\Lambda_2) \}.
\end{align*}
Then we obtain
\begin{align*}
{\mathcal F} :=& \  \frac{1}{n} \| {\bm y}^* - X V_1 {\bm \beta} \|_2^2 + \frac{\rho_2}{2} \left\| V_1 - V_0 + \Lambda_2 \right\|_F^2 \\
=& \ \frac{1}{n} {\bm \beta}^T V_1^T X^T X V_1 {\bm \beta} - \frac{2}{n} {\rm tr} ({\bm \beta} {\bm y}^{*T} X V_1) + \frac{\rho_2}{2} {\rm tr} (V_1^T V_1) - \rho_2 {\rm tr} \{ (V_0-\Lambda_2)^T V_1\} + C,
\end{align*}
where $C$ is a constant.
By $\partial {\mathcal F} / \partial V_1 = {\bm O}$, we have
\begin{equation*}
\frac{2}{n} X^T X V_1 {\bm \beta} {\bm \beta}^T - \frac{2}{n} X^T {\bm y}^* {\bm \beta}^T + \rho_2 V_1 - \rho_2 (V_0 - \Lambda_2) = \bm O.
\end{equation*}
This leads to the update of $V_1$. 
%\begin{equation*}
%{\rm vec} (V_1) = \left( \frac{1}{n} {\bm \beta} {\bm \beta}^T \otimes X^T X + \frac{\rho_2}{2} I_k \otimes I_p \right)^{-1} {\rm vec} \left\{ \frac{1}{n} X^T {\bm y} {\bm \beta}^T + \frac{\rho_2}{2} (V_0 - \Lambda_2) \right\}.
%\end{equation*}
%%%
\item[Update of $V$.]
\begin{equation*}
V := \argmin_{V} \left\{ \frac{w}{n} \| X - Z V^T \|_F^2 + \frac{\rho_1}{2} \| V - V_0 + \Lambda_1 \|_F^2 \right\} \ \ {\rm subject \ to} \ \ V^T V = I_k.
\end{equation*}
The terms of the right-hand side are, respectively, calculated as
\begin{align*}
\| X - Z V^T \|_F^2 &= {\rm tr} (X^T X) - 2 {\rm tr} (V Z^T X) + {\rm tr}(Z^T Z), \\
\| V - V_0 + \Lambda_1 \|_F^2 &= -2 {\rm tr} \{ (V_0-\Lambda_1)^T V\} + {\rm tr}\{ (V_0-\Lambda_1)^T (V_0-\Lambda_1) \} + k.
\end{align*}
With the equality constraint $V^T V = I_k$, we get
\begin{align*}
&\argmin_{V} \left\{ \frac{w}{n} \| X - Z V^T \|_F^2 + \frac{\rho_1}{2} \| V - V_0 + \Lambda_1 \|_F^2 \right\} \\
=& \argmin_{V} \left\{ \left\| V - \left\{ \frac{w}{n} X^T Z + \frac{\rho_1}{2} (V_0 - \Lambda_1) \right\} \right\|_F^2 \right\}.
\end{align*}
By the SVD ${w} X^T Z/n + {\rho_1}\left(V_0 - \Lambda_1 \right)/2 = P \Omega Q^T$, we have the solution $V = P Q^T$. 
This follows the Procrustes rotation by Zou \textit{et al.} (2006). 
%%%
\item[Update of $V_0$.]
\begin{equation}
V_0 := \argmin_{V_0} \left\{ \frac{\rho_1}{2} \| V - V_0 + \Lambda_1 \|_F^2 + \frac{\rho_2}{2} \| V_1 - V_0 + \Lambda_2 \|_F^2 + \lambda_V \| V_0 \|_1 \right\}.
\label{eq:updateV0_ADMM}
\end{equation}
By a simple calculation, the first two terms of the right-hand side are calculated by
\begin{equation*}
\frac{\rho_1+\rho_2}{2} \left\| V_0 - \frac{1}{\rho_1+\rho_2} \{ \rho_1 (V+\Lambda_1) + \rho_2 (V_1+\Lambda_2) \} \right\|_F^2.
\end{equation*}
Formula \eqref{eq:updateV0_ADMM} is rewritten by
\begin{equation*}
V_0 := \argmin_{V_0} \left\{ \frac{1}{2} \left\| V_0 - \frac{1}{\rho_1+\rho_2} \left\{ \rho_1 (V+\Lambda_1) + \rho_2 (V_1+\Lambda_2) \right\} \right\|_F^2 + \frac{\lambda_V}{\rho_1+\rho_2} \| V_0 \|_1 \right\}.
\end{equation*}
Thus, we obtain the update of $V_0$. 
%\begin{equation}
%v_{0ij} = {\mathcal S} \left( \frac{\rho_1(v_{ij} + \lambda_{1ij}) +\rho_2(v_{ij} + \lambda_{2ij})}{\rho_1 + \rho_2}, \frac{\lambda_V}{\rho_1 + \rho_2} \right), \quad i=1,\ldots,p, \ j=1,\ldots,k.
%\label{eqn:V0}
%\end{equation}
%Here ${\mathcal S}$ is the soft-thresholding operator and $\lambda_{kij} \ (k=1,2)$ is the $(i,j)$-th element of the matrix $\Lambda_k \ (k=1,2)$. 
%%%
\item[Update of $Z$.]
\begin{equation*}
Z := \argmin_{Z}  \left\{ \frac{w}{n} \| X - Z V^T \|_F^2 \right\}.
\end{equation*}
We have the solution $Z=X V$ from the first order optimality condition.  
%%%
\item[Update of $\bm \beta$.]
\begin{equation*}
{\bm \beta} := \argmin_{\bm \beta} \left\{ \frac{1}{n} \| {\bm y} - \beta_0 {\bm 1}_n - X V_1 {\bm \beta} \|_2^2 + \frac{\rho_2}{2} \| {\bm \beta} - {\bm \beta}_0 + {\bm \lambda} \|_2^2 \right\}.
\end{equation*}
The first order optimality condition leads to
\begin{equation*}
- \frac{2}{n} V_1^T X^T ({\bm y} - \beta_0 {\bm 1}_n - X V_1 {\bm \beta}) + \rho_2 ({\bm \beta} - {\bm \beta}_0 + {\bm \lambda}) = {\bm 0}.
\end{equation*}
This leads to the update of $\bm \beta$. 
%\begin{equation*}
%\bm \beta = \left( \frac{1}{n} V_1^T X^T X V_1 + \frac{\rho_2}{2} I_k \right)^{-1} \left\{ \frac{1}{n} V_1^T X^T {\bm y} + \frac{\rho_2}{2} ({\bm \beta}_0 - {\bm \lambda}) \right\}.
%\end{equation*}
%%%
\item[Update of $\bm \beta_0$.]
\begin{equation*}
{\bm \beta}_0 := \argmin_{{\bm \beta}_0} \left\{  \frac{\rho_2}{2} \| {\bm \beta} - {\bm \beta}_0 + {\bm \lambda} \|_2^2 + \lambda_\beta \| {\bm \beta}_0 \|_1\right\}.
\end{equation*}
It is clear that the update of ${\bm \beta}_0$ is simply obtained by element-wise soft-threshold operator.
%\begin{equation*}
%\beta_{0j} = {\mathcal S} \left( \beta_j + \lambda_{j}, \frac{\lambda_\beta}{\rho_2} \right), \quad j=1,\ldots,k.
%\end{equation*}
\end{description}

%%%%%%%%%%%%%%%%%%%%%%%%%%%%%%%%%%%%%%%%%%%%%%%%%%%%%%%%%%%%%%%%%%%%%%%%%%%%%%%%%%
\renewcommand{\theequation}{B.\arabic{equation}}
\setcounter{equation}{0}
\subsection*{B \quad The LADMM algorithm for SPCRsvd}
%\label{sec:AppendixB}
The LADMM algorithm for SPCRsvd is given as follows:
\begin{description}
%%%
\item[Step 1] Set the values of the tuning parameter $w$, the regularization parameters $\lambda_V, \lambda_{\bm \beta}$, and the penalty parameters $\rho_1, \rho_2$. 
%%%
\item[Step 2] Initialize the all parameters by $\beta_0^{(0)}, {\bm \beta}^{(0)}, {\bm \beta}_0^{(0)}, Z^{(0)}, V^{(0)}, V_0^{(0)}, \Lambda^{(0)},{\bm \lambda}^{(0)}$. 
%%%
\item[Step 3] For $m=0,1,2,\ldots$, repeat from Step 4 to Step 10 until convergence.
%%%
\item[Step 4] Update $V$ as follows:
\begin{equation*}
V^{(m+1)}=PQ^T,
\end{equation*}
where $P$ and $Q$ are the matrices given by the SVD 
\begin{equation*}
\frac{w}{n} X^T Z^{(m)} + \frac{\rho_1}{2} \left(V_0^{(m)} + \Lambda^{(m)} \right) = P \Omega Q^T.
\end{equation*}
%%%ここ変更する
\item[Step 5] Update $V_0$ as follows:
\begin{equation}
v_{0ij}^{(m+1)} = {\mathcal S} \left( s_{ij}, \lambda_V/\left( \frac{2 \nu + n \rho_1}{n} \right) \right), \quad i=1,\ldots,p, \ j=1,\ldots,k,
\label{eq:V0_LADMM}
\end{equation}
where $v_{0ij}^{(m)}=(V_0^{(m)})_{ij}$, $\nu$ is the maximum eigenvalue of ${\bm \beta}^{(m)} {\bm \beta}^{(m)T} \otimes X^T X$, and $s_{ij}$ is the $(i,j)$-th element of the matrix
\begin{equation*}
\frac{2n}{2 \nu + n \rho_1} \left\{\frac{1}{n} \left( X^T ({\bm y} - \beta_0^{(m)} {\bm 1}_n) {\bm \beta}^{(m)T} - X^T X V_0^{(m)} {\bm \beta}^{(m)} {\bm \beta}^{(m)T} ) + \frac{\nu}{n} V_0^{(m)} - \frac{\rho_1}{2} ( \Lambda^{(m)} - V^{(m+1)} \right) \right\}.
\end{equation*}
%\begin{equation*}
%v_{0ij}^{(m+1)} = {\mathcal S} \left( \frac{\rho_1(v_{ij}^{(m+1)} + \lambda_{1ij}^{(m)}) +\rho_2(v_{ij}^{(m+1)} + \lambda_{2ij}^{(m)})}{\rho_1 + \rho_2}, \frac{\lambda_V}{\rho_1 + \rho_2} \right), \quad i=1,\ldots,p, \ j=1,\ldots,k,
%%\label{eqn:V0}
%\end{equation*}
%where $v_{0ij}^{(m)}=(V_0^{(m)})_{ij}$, $v_{ij}^{(m)}=(V^{(m)})_{ij}$, $\lambda_{\ell ij} \ (\ell =1,2)$ is the $(i,j)$-th element of the matrix $\Lambda_\ell \ (\ell=1,2)$, and ${\mathcal S} (\cdot,\cdot) $ is the soft-thresholding operator defined by ${\mathcal S}(x,\lambda)={\rm sign}(x)(|x|-\lambda)_+$.
%%%
\item[Step 6] Update $Z$ by $Z^{(m+1)}=X V^{(m+1)}$.
%%%
\item[Step 7] Update ${\bm \beta}$ as follows:
\begin{equation*}
\bm \beta^{(m+1)} = \left(\frac{1}{n} V_0^{(m+1)T} X^T X V_0^{(m+1)} + \frac{\rho_2}{2} I_k \right)^{-1} \left\{ \frac{1}{n} V_0^{(m+1)T} X^T ({\bm y} - \beta_0^{(m)} {\bm 1}_n) + \frac{\rho_2}{2} ({\bm \beta}_0^{(m)} - {\bm \lambda}^{(m)})  \right\}.
\end{equation*}
%%%
\item[Step 8] Update ${\bm \beta}_0$ as follows:
\begin{equation*}
\beta_{0j}^{(m+1)} = {\mathcal S} \left( \beta_j^{(m+1)} + \lambda_{j}^{(m)}, \frac{\lambda_\beta}{\rho_2} \right), \quad j=1,\ldots,k,
%\label{eqn:beta0}
\end{equation*}
where $\lambda_{j}^{(m)}$ and $\beta_j^{(m)}$ are the $j$-th element of the vector ${\bm \lambda}^{(m)}$ and ${\bm \beta}^{(m)}$, respectively. 
%%%
\item[Step 9] Update $\beta_0$ as follows:
\begin{equation*}
\beta_{0}^{(m+1)} = \frac{1}{n} {\bm 1}_n ^T ( {\bm y} - X V_0^{(m+1)} {\bm \beta}^{(m+1)} )
%\label{eqn:beta0}
\end{equation*}
%%%
\item[Step 10] Update ${\Lambda},{\bm \lambda}$ as follows:
\begin{align*}
\Lambda^{(m+1)} &= \Lambda^{(m)} + V_0^{(m+1)} - V^{(m+1)},\\
{\bm \lambda}^{(m+1)} &= {\bm \lambda}^{(m)} + {\bm \beta}^{(m+1)} - {\bm \beta}^{(m+1)}_0.
\end{align*}
\end{description}

Next, we describe the update of only $V_0$, because the derivations of other updates are same with Appendix A.
Similar with Appendix A, we omit the index $m$ for iteration. 

We consider
\begin{equation}
V_0 := \argmin_{V_0} \left\{ \frac{1}{n} \| {\bm y} - \beta_0 {\bm 1}_n - X V_0 {\bm \beta} \|_2^2 + \frac{\rho_1}{2} \| V_0 - V+ \Lambda \|_F^2 + \lambda_V \| V_0 \|_1 \right\}.
\label{eq:updateV}
\end{equation}
Set ${\bm y}^* = {\bm y} - \beta_0 {\bm 1}_n$. 
By Taylor expansion, the term $\| {\bm y}^* - X V_0 {\bm \beta} \|_2^2$ is approximated as
\begin{align*}
\| {\bm y}^* - X V_0 {\bm \beta} \|_2^2 &= {\bm y}^{*T} {\bm y}^* - 2 {\rm tr} ( {\bm \beta} {\bm y}^{*T} X V_0 ) + {\bm \beta}^T V_0^T X^T X V_0 {\bm \beta} \nonumber \\
&\approx {\bm y}^{*T} {\bm y}^* - 2 {\rm tr} ( {\bm \beta} {\bm y}^{*T} X V_0 ) + 2 {\rm tr} ( {\bm \beta} {\bm \beta}^T \tilde{V}_0 X^T X V_0 ) + \nu \| V_0 - \tilde{V}_0 \|_F^2,
\end{align*}
where $\tilde{V}_0$ is the current estimate of $V_0$ and $\nu$ is a constant. 
According to Li \textit{et al.} (2014), we use the maximum eigenvalue of ${\bm \beta} {\bm \beta}^T \otimes X^T X$ as $\nu$. 
Using the approximation, the problem \eqref{eq:updateV} can be replaced with 
\begin{align*}
V_0 &:= \argmin_{V_0} \bigg\{  \underbrace{- \frac{2}{n} {\rm tr} ( {\bm \beta} {\bm y}^{*T} X V_0 ) + \frac{2}{n} {\rm tr} ( {\bm \beta} {\bm \beta}^T \tilde{V}_0 X^T X V_0 ) + \frac{\nu}{n} \| V_0 - \tilde{V}_0 \|_F^2 + \frac{\rho_1}{2} \| V_0 - V + \Lambda \|_F^2}_{\rm (A)} \\
&+ \lambda_V \| V_0 \|_1 \bigg\}.
\end{align*}
Formula (A) is calculated as
\begin{equation*}
\frac{2 \nu + n \rho_1}{2n} \left\| V_0 - \frac{2n}{2 \nu + n \rho_1} \left\{ \frac{1}{n} (X^T {\bm y}^* {\bm \beta}^T - X^T X \tilde{V}_0 {\bm \beta} {\bm \beta}^T) + \frac{\nu}{n} \tilde{V}_0 - \frac{\rho_1}{2} (\Lambda - V) \right\} \right\|_F^2.
\end{equation*}
This leads to the update of $V_0$ in Formula \eqref{eq:V0_LADMM}.

%%%%%%%%%%%%%%%%%%%%%%%%%%%%%%%%%%%%%%%%%%%%%%%%%%%%%%%%%%%%%%%%%%%%%%%%%%%%%%%%%%%%%%%%%%%%%%%%%%%%%%%%%%%%%%%%%%%%%%%%%%%%%%%%%%%%%%%%%%%%%%%%%%%%%%%%%%%%%%%%
\section*{Acknowledgements}
%The authors thank the reviewers for their helpful comments and constructive suggestions. 
%This work was supported by JSPS KAKENHI Grant Number JP19K11854 and MEXT KAKENHI Grant Numbers JP16H06429, JP16K21723, and JP16H06430. 
The author was supported by JSPS KAKENHI Grant Number JP19K11854 and MEXT KAKENHI Grant Numbers JP16H06429, JP16K21723, and JP16H06430. 
%The computational resource was provided by the Super Computer System, Human Genome Center, Institute of Medical Science, The University of Tokyo.

%%%%%%%%%%%%%%%%%%%%%%%%%%%%%%%%%%%%%%%%%%%%%%%%%%%%%%%%%%%%%%%%%%%%%%%%%%%%%%%%%%%%%%%%%%%%%%%%%%%%%%%%%%%%%%%%%%%%%%%%%%%%%%%%%%%%%%%%%%%%%%%%%%%%%%%%%%%%%%%%

%\bibliographystyle{imsart-number} % Style BST file (imsart-number.bst or imsart-nameyear.bst)
%\bibliography{KawanoEtAl}       % Bibliography file (usually '*.bib')

%\bibliographystyle{apalike}
%\bibliography{KawanoEtAl}

\end{document}